\documentclass{article}
\usepackage[final]{nips_2018}

\usepackage[utf8]{inputenc} % allow utf-8 input
\usepackage[T1]{fontenc}    % use 8-bit T1 fonts
\usepackage{hyperref}       % hyperlinks
\usepackage{url}            % simple URL typesetting
\usepackage{booktabs}       % professional-quality tables
\usepackage{amsfonts}       % blackboard math symbols
\usepackage{nicefrac}       % compact symbols for 1/2, etc.
\usepackage{microtype}      % microtypography

\usepackage{amsmath}
\usepackage{multirow}
\usepackage{graphicx}
\usepackage{subcaption}
\DeclareGraphicsExtensions{.eps,.pdf,.jpeg,.png}

\title{AffinityNet: Semi-supervised Few-shot Learning for Disease Type Prediction}

\author{
Tianle Ma\\
  Department of Computer Science and Engineering\\
  University at Buffalo\\
  Buffalo, NY 14260 \\
  \texttt{tianlema@buffalo.edu}\\
  \AND
Aidong Zhang\\
Department of Computer Science and Engineering\\
University at Buffalo\\
Buffalo, NY 14260 \\
\texttt{azhang@buffalo.edu}
}

\begin{document}

\maketitle

\begin{abstract}
While deep learning has achieved great success in computer vision and many other fields, currently it does not work very well on patient genomic data with the ``big $p$, small $N$'' problem (i.e., a relatively small number of samples with high-dimensional features). In order to make deep learning work with a small amount of training data, we have to design new models that facilitate few-shot learning. Here we present the Affinity Network Model (AffinityNet), a data efficient deep learning model that can learn from a limited number of training examples and generalize well. The backbone of the AffinityNet model consists of stacked k-Nearest-Neighbor (kNN) attention pooling layers. The kNN attention pooling layer is a generalization of the Graph Attention Model (GAM), and can be applied to not only graphs but also any set of objects regardless of whether a graph is given or not. As a new deep learning module, kNN attention pooling layers can be plugged into any neural network model just like convolutional layers. As a simple special case of kNN attention pooling layer, feature attention layer can directly select important features that are useful for classification tasks. Experiments on both synthetic data and cancer genomic data from TCGA projects show that our AffinityNet model has better generalization power than conventional neural network models with little training data.
We have implemented our method using PyTorch framework (\href{https://pytorch.org}{https://pytorch.org}).The code is freely available at \href{https://github.com/BeautyOfWeb/AffinityNet}{https://github.com/BeautyOfWeb/AffinityNet}.
\end{abstract}
	
\maketitle
	
\section{Introduction}
Patients, drugs, networks, etc., are all complex objects with heterogeneous features or attributes. Complex object clustering and classification are ubiquitous in real world applications. For instance, it is important to cluster cancer patients into subgroups and identify disease subtypes in cancer genomics \citep{Shen2012,Wang2014,Ma2017}. Compared with images, which have homogeneous structured features (i.e., pixels are arranged in a 3-D array as raw features), complex objects usually have heterogeneous features with unclear structures. Deep learning models such as Convolutional Neural Networks (CNNs) widely used in computer vision \citep{LeCun2015,Krizhevsky2012} and other fields \citep{Bahdanau2014,Sutskever2014,Silver2016,Banino2018} cannot be directly applied to complex objects whose features are not ordered structurally.
	
One critical challenge in cancer patient clustering problem is the ``big $p$, small $N$'' problem: we only have a relatively small number of samples (i.e., patients) compared with high-dimensional features each sample has. In other words, we do not have an ``ImageNet''\citep{Russakovsky2015} to train deep learning models that can learn good representations from raw features. Moreover, unlike pixels in images, patient features such as gene expressions are much noisier and more heterogeneous. These features are not ``naturally'' ordered. Thus we cannot directly use convolutional neural networks with small filters to extract abstract local features.

For a clustering/classification task, nodes/objects belonging to the same cluster should have similar representations that are near the cluster centroid. Based on this intuition we developed the \textbf{k-nearest-neighbor (kNN) attention pooling} layer, which applies the attention mechanism to learning node representations. With the kNN attention pooling layer, each node's representation is decided by its k-nearest neighbors as well as itself, ensuring that similar nodes will have similar learned representations. Similar to Graph Attention Model (GAM) \citep{Velickovic2017}, we propose the Affinity Network Model (AffinityNet) that consists of stacked kNN attention pooling layers to learn the deep representations of a set of objects.
While GAM is designed to tackle representation learning on graphs \citep{Velickovic2017, Hamilton2017a} and it does not directly apply to data without a known graph, our AffinityNet model generalizes GAM to facilitate representation learning on any collections of objects with or without a known graph.

In addition to learning deep representations for classifying objects, feature selection is also important in biomedical research. Though the number of features (i.e., variables or covariates) in genomic data is usually very high, many features may be irrelevant to a specific task. For instance, a disease may only have a few risk factors involving a small number of features. In order to facilitate feature selection in a ``deep learning'' way, we propose a \textbf{feature attention} layer, a simple special case of the kNN attention pooling layer which can be incorporated into a neural network model and directly learn feature weights using backpropagation.
	
We performed experiments on both synthetic and real cancer genomics data. The results demonstrated that our AffinityNet model has better generalization power than conventional neural network models for few-shot learning.
	
\section{Related work}	
kNN attention pooling layer is related to graph learning \citep{Hamilton2017a,Kipf2016,Velickovic2017}, attention model \citep{Vaswani2017,Velickovic2017}, pooling and normalization layers \citep{Ioffe2015} in deep learning literature. 
	
In graph learning, a graph has a number of nodes and edges (both nodes and edges can have features). When available, combining node features with graph structure can do a better job than using node features alone. For example, Graph Convolutional Neural Network \citep{Kipf2016} incorporates graph structure (i.e. edges) into the learning process to facilitate semi-supervised few-shot learning.
Graph Attention Model (GAM) \citep{Velickovic2017} learns a representation for each node based on the weighted pooling (i.e., attention) of its neighborhood in the given graph, and then performs classification using the learned representations.
However, all these graph learning algorithms require that a graph (i.e., edges among nodes) is known. Many algorithms also require the input to be the whole graph \citep{Velickovic2017}, and thus do not scale well to large graphs.
Our proposed AffinityNet model generalizes graph learning to a collections of objects (e.g., patients) with or without known graphs. 
	
As the key component of AffinityNet, kNN attention pooling layer is also related to normalization layers in deep learning, such as batch normalization \citep{Ioffe2015}, instance normalization \citep{Jing2017}, or layer normalization \citep{Ba2016}. 
%In batch normalization \citep{Ioffe2015}, a batch is used to calculate mean and variance and perform normalization. In test time, the overall mean and variance will be used to normalize samples. Batch size is an important parameters. During training, batch size must be bigger than one.
%Layer normalization \citep{Ba2016} gets rid of batch size constraints, instead to make each layer normalized. Instance normalization is similar. 
All these normalization layers use batch statistics or feature statistics to normalize instance features, while kNN attention pooling layers apply the attention mechanism to the learned instance representations to ensure similar instances having similar representations.
	
The kNN attention pooling layer is different from the existing max or average pooling layers used in deep learning models, where features in a local neighborhood are pooled to extract the signal and reduce feature dimensions. Our proposed kNN attention pooling layer applies pooling on node representations instead of individual features.
The kNN attention pooling layer combines normalization, attention and pooling, making it more general and powerful. It can serve as an implicit regularizer to make the network generalize well for semi-supervised few-shot learning.

\section{Affinity Network Model (AffinityNet)}
One key ingredient for the success of deep learning is its ability to learn a good representation \citep{Bengio2013} through multiple complex nonlinear transformations. For classification tasks, the learned representation (usually the last hidden layer) is often linearly separable for different classes. If the output layer is a fully connected layer for classification, then the weight matrix for the last layer can be seen as the class centroids in the transformed feature space.
While conventional deep learning models often perform well when lots of training data is available, our goal is to design new models that can learn a good feature transformation in a transparent and data efficient way. 
	
\begin{figure}[!t]
		\centering
		\includegraphics[width=.8\textwidth]{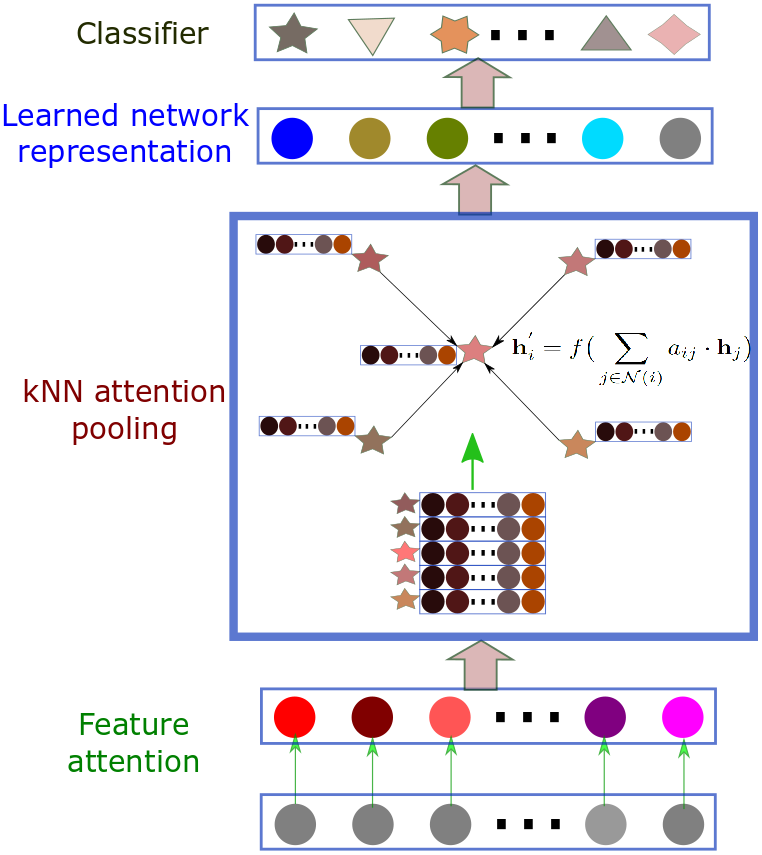}
		\caption{AffinityNet model overview}
		\label{fig:model-overview}
	\end{figure}

Built upon the existing modules in the deep learning toolbox, we developed the kNN attention pooling layer, and used it to construct the AffinityNet Model. In a typical AffinityNet model as shown in Fig.~\ref{fig:model-overview},  the input layer is followed by a feature attention layer (a simple special case of kNN attention pooling layer used for raw feature selection), and then followed by multiple stacked kNN attention pooling layers (Fig.~\ref{fig:model-overview} only illustrates one kNN attention pooling layer). 
The output of the last kNN attention pooling layer will be the new learned network representations, which can be used for classification (as depicted in Fig.~\ref{fig:model-overview}) or regression tasks (for example, Cox model \citep{Mobadersany2018}). 
Though it is possible to train AffinityNet with only a few labeled examples, it is more advantageous to use it as a semi-supervised learning framework (i.e., using both labeled and unlabeled data during training).
	
As the main components of AffinityNet are stacked kNN attention pooling layers, we describe it in detail in the following section.

\subsection{kNN attention pooling layer}
A good classification model should have the ability to learn a feature transformation such that objects belonging to the same class have similar representations which are near the class centroid in the transformed feature space.
	
	%In the beginning, class centroids are unknown, how to find them automatically? (kMean clustering provides a way find cluster centroids using EM algorithm if we already have a good feature representation.) 
	%We propose a k-nearest-neighborhood (kNN) pooling layer that utilizes attention mechanism to calculate a weighted representation that incorporates the local neighborhood information.
	
As an object's k-nearest neighbors should have similar feature representations, we propose the kNN attention pooling layer to incorporate neighborhood information using attention-based pooling (Eq.~\ref{eq:knn_pooling}): 
	
	\begin{equation} \label{eq:knn_pooling}
	\mathbf{h}^{'}_{i} = f\big(\sum_{j \in \mathcal{N}(i)}a(\mathbf{h}_i, \mathbf{h}_j) \cdot \mathbf{h}_j\big)
	\end{equation}
	
In Eq.~\ref{eq:knn_pooling}, $\mathbf{h}_i$ and $\mathbf{h}^{'}_{i}$ are input feature representations and transformed feature representations for object $i$, respectively.  $\mathcal{N}(i)$ represents the neighborhood of object $i$. If a graph is given as in the graph learning setting \citep{Hamilton2017a}, we can use the given graph to determine the neighborhood. If the given graph is very large with a high degree, in order to reduce the computational cost, we can randomly sample $k$ ($k$ is a fixed small number) neighbors for computing Eq.~\ref{eq:knn_pooling} \citep{Hamilton2017}. In the kNN attention pooling layer, $k$ is a hyperparameter that determines how many neighbors are used for calculating the representation of a node. 
	
$f(\cdot)$ is a nonlinear transformation, for example, an \textit{affine} layer with weight $\mathbf{W}$ and bias $\mathbf{b}$ followed by \textit{ReLU()} nonlinear activation:

\begin{equation}\label{eq:f}
	f(\mathbf{h}) = max(\mathbf{Wh + b, 0})
\end{equation}

$a_{ij} = a(\mathbf{h}_i, \mathbf{h}_j)$ in Eq.~\ref{eq:knn_pooling} is the normalized attention from object $i$ to object $j$.  $a(\mathbf{\cdot}, \mathbf{\cdot})$ is the attention kernel that will be discussed in the next section.
	
\subsubsection{Attention kernels} \label{sec:attention_kernel}
Intuitively, if two objects are similar, their feature representations should be near each other. Objects belonging to the same class should be clustered together in the learned feature space. In order to achieve this, kNN attention pooling layer uses weighted pooling to ``attract'' similar objects together in the transformed feature space. Attention kernels essentially calculate the similarities among objects to facilitate weighted pooling. 
	
There are many choices of attention kernels. For example: 
	
	\begin{itemize}
		\item {Cosine similarity:
			\begin{equation}
			\alpha_{ij} =  \frac{\mathbf{h}_i \cdot \mathbf{h}_j}{||\mathbf{h}_i||\cdot ||\mathbf{h}_j||}
			\end{equation}
		}
		
		\item {Inner product \citep{Vaswani2017}:
			\begin{equation}
			\alpha_{ij} =  \mathbf{h}_i \cdot \mathbf{h}_j
			\end{equation}
		}
		
		\item {Perceptron affine kernel \citep{Velickovic2017}:
			\begin{equation}
			\alpha_{ij} = \mathbf{w}^T \cdot (\mathbf{h}_i || \mathbf{h}_j)
			\end{equation}
		}
		
		\item {Inverse distance with weighted $L_2$ norm ($\mathbf{w}$ is the feature weight):
			\begin{equation}
			\begin{split}
			\alpha_{ij} &= -||\mathbf{w} \odot \mathbf{h}_i - \mathbf{w} \odot \mathbf{h}_j||^2\\
			\end{split}
			\end{equation}
		}
	\end{itemize}
	
In order to calculate a weighted average of new representations, we can use the \textit{Softmax} function to normalize the attention (other normalization is also feasible). Therefore the normalized attention kernel is: 
	
	\begin{equation}\label{eq:weighted_attention}
	a_{ij} = a(\mathbf{h}_i, \mathbf{h}_j) = \frac{e^{\alpha_{ij}}}{\sum_{j \in \mathcal{N}(i)} e^{\alpha_{ij}}}
	\end{equation}
	
Now $\sum_{j \in \mathcal{N}(i)} a_{ij} = 1, a_{ij}\ge 0$. Note for each node $i$, we only select its neighbors $\mathcal{N}(i)$ for normalization. 
If the graph is not given, in order to determine $\mathcal{N}(i)$, we can use attention kernel to calculate an affinity/similarity graph (i.e., the similarities among all the objects), and then use this affinity graph to decide the neighborhood $\mathcal{N}(i)$. As an additional regularizer, we can use one type of affinity kernels to calculate the affinity graph and another to compute the normalized attention.
	
\subsubsection{Layer-specific dynamic affinity graph}
The kNN attention pooling layer can be applied to a collection of objects regardless of whether a graph (e.g., links among objects) is given or not. 
If a graph is given, we can directly use the graph to determine the neighborhood in Eq.~\ref{eq:knn_pooling} and Eq.~\ref{eq:weighted_attention}, which is the same as in Graph Attention Model \citep{Velickovic2017}. If the degree of the graph is too high, and some nodes have very large neighborhoods, then we can select only $k$ nearest neighbors for calculating the attention when the computational cost is a big concern. 
	
Regardless of whether a graph is given or not, we can always calculate an affinity graph $\mathbf{G}_n$ based on node features using some similarity metric including the attention kernels discussed in Sec.~\ref{sec:attention_kernel}. 
As our AffinityNet model contains multiple kNN pooling layers stacked together, we can calculate a layer-specific dynamic affinity graph using the learned node feature representations from each layer during training.
	
Also, we can use the graph calculated using features from the previous layer to determine the k-nearest-neighborhood for the next layer. This can be seen as an implicit regularizer preventing the learned representation from drifting away from the previous layer too much in a single layer operation. 
Mathematically, for layer $l$, we can calculate a layer-specific dynamic affinity graph $\mathbf{G}^{(l)}$ using Eq.~\ref{eq:dynamic_graph}.
	
	\begin{equation}\label{eq:dynamic_graph}
	\mathbf{G}^{(l)} = \lambda \mathbf{G}_e + (1-\lambda) \big( \eta \mathbf{G}_{n}^{(l)} + (1-\eta) \mathbf{G}_{n}^{(l-1)} \big)
	\end{equation}
	
In Eq.~\ref{eq:dynamic_graph}, $\mathbf{G}_e$ is the given graph if available. When not available, we can simply set $\lambda=0 (0\le \lambda \le 1)$. $\mathbf{G}_n^{(l)}$ and $\mathbf{G}_n^{(l-1)}$ are the node-feature-derived affinity graphs for the current layer $l$ and the previous layer $l-1$, respectively. We can combine $\mathbf{G}_n^{(l-1)}$ and $\mathbf{G}_n^{(l)}$ with a parameter $\eta, 0\le \eta \le 1$. If $\eta=0$ in Eq.~\ref{eq:dynamic_graph}, then only $\mathbf{G}_n^{(l-1)}$ is used; if $\eta=1$, then only $\mathbf{G}_n^{(l)}$ is used. In practice, we can set $\eta=0.5$.
	
If the input of the AffinityNet model consists of $N$ objects, then we will learn dynamic affinity graphs for these $N$ objects during training. After training, the final learned affinity graph from the last layer can also be used for spectral clustering (affinity graphs calculated using higher-level features may be more informative for separating different classes). In this sense, we also call our framework affinity network learning.
	
\subsubsection{Semi-supervised few-shot learning}
Semi-supervised few-shot learning \citep{Ravi2016,Kingma2014,Kipf2016,Rasmus2015} only allows using very few labeled instances to train a model and requires the model to generalize well. Our proposed AffinityNet model consisting of stacked kNN pooling layers can perform a good job towards this goal.
More specifically, for cancer patient clustering problems, we usually have several hundred patients in a study. If we can obtain a few labeled training examples (for example, human experts can manually assign labels for some patients), we can use the AffinityNet model for semi-supervised learning. The input of the AffinityNet model is the patient-feature matrix consisting of all patients, and the output of the model is the newly learned patient representations as well as class labels. We only backpropagate the classification error for those labeled patients. Different from conventional neural network models where each instance is independently trained without batch normalization \citep{Ioffe2015}, AffinityNet can utilize unlabeled instances for calculating kNN attention-based representations in the whole sample pool. 
In a sense, the kNN attention pooling layer performs both nonlinear transformation and ``clustering'' (attracting similar instances together in the learned feature space) during training. Even though the labels of most patients are unknown, their feature representations can be used for learning a global affinity graph, which is useful to cluster or classify all patients in the cohort.
Our AffinityNet model can also be used for data distillation \citep{radosavovic2017data}. We can train a few examples with true labels, and use our learned model to generate some noisy labels for unlabeled data. Then we can train our model with both clean and noisy labels and repeat this process iteratively. 

When dealing with very large graphs, we can feed a small batch of instances (i.e., a partial graph) at a time to the AffinityNet model to reduce the computational burden. Though each batch may contain different instances, the kNN pooling layer can still work well with the attention mechanism. Our PyTorch implementation of AffinityNet can even handle the extreme case where only one instance is fed into the model at a time, in which case the AffinityNet model operates as conventional deep learning model to only learn a nonlinear transformation without kNN attention pooling operation.
	
%\subsection{Multi-view fusion}
%If multi-view data is available, we can use perform multi-view learning. The simplest approach is to concatenate all views and treat features from multi-view the same.  
	%
	%However, we hypothesize it is better to process each view separately and learn an higher-level representation for each view and then combine them later.
	
	%\subsubsection{View fusion layer}
	%After learning a high-level representation for each view separately, we can merge them to get an integrated view. 
	%
	%Learned high-level representation for each view separately:
	%\begin{equation}
	%\mathbf{h}_i^{'}=(\mathbf{h}_i^{(1)} || \mathbf{h}_i^{(2)} || \cdots || \mathbf{h}_i^{(V)})
	%\end{equation}
	
	%The simplest approach is to calculate a weighted view. For this we must have the same dimension for each view.
	%\begin{equation}
	%\mathbf{h}_i^{'} = \sum_{v=1}^{V} \beta_{v} \mathbf{h}_i^{(v)}
	%\end{equation}
	%
	%More generally, we can concatenate the learned views in higher layer and use a fully connected layer to ``merge or fuse'' multi-views into a single vector representation. 

	%In most cases, we might collect a lot of useless features. The Euclidean distance using all features will be skrewed from truth. We propose to use weighted Euclidean distance:
	%
	%\begin{equation}
	%d_{ij}^{'} = ||\mathbf{h}^{'}_{i}-\mathbf{h}^{'}_{j}|| =||\mathbf{w} \odot \mathbf{h}_i - \mathbf{w} \odot \mathbf{h}_j||^2
	%\end{equation}
	
\subsection{Feature Attention Layer}
Deep neural networks can learn good hierarchical local feature extractors (such as convolutional filters or inception modules \citep{Szegedy2017}) automatically through gradient descent. Local feature operations such as convolutions require features to be ordered structurally. For images or videos, pixels near each other naturally form a neighborhood. However, in other applications, features are not ordered and the structural relations among features are unknown. Therefore we cannot directly learn a local feature extractor, instead we have to learn a feature selector that can select important individual features.

In addition, there can be many redundant, noisy, or irrelevant features, and the Euclidean distance between objects using all the features may be dominated by the irrelevant ones \citep{Bellet2013}. However, with proper feature weighting, we can separate objects from different classes well. This motivates us to develop a feature attention layer as a simple special case of kNN attention pooling layer.

Let $\mathbf{h}_i \in \mathbb{R}^{p}$ be the feature vector of object $i$, and $\mathbf{w} \in \mathbb{R}^{p}$ be the feature attention (i.e., weight), satisfying

\begin{equation} \label{eq:w}
\mathbf{w}=(w_1, w_2, \cdots, w_p), \quad \sum_{j=1}^{p}w_j = 1, w_j \ge 0
\end{equation}

Instead of the commonly used \textit{affine} transformation followed by \textit{ReLU()} nonlinearity as in Eq.~\ref{eq:f}, the feature attention layer performs element-wise multiplication (Eq.~\ref{eq:feature_attention}, $\odot$ is element-wise multiplication operator) with the weight constraint (Eq.~\ref{eq:w}). This is the only difference between the feature attention layer and the kNN attention pooling layer.

\begin{equation} \label{eq:feature_attention}
f(\mathbf{h}_{i}) = \mathbf{w} \odot \mathbf{h}_{i}
\end{equation}

\begin{equation}\label{eq:d_ij}
d_{ij} = ||\mathbf{h}_{i}-\mathbf{h}_{j}||
\end{equation}

\begin{equation}\label{eq:d_ij_weighted}
d_{ij}^{'} = ||f(\mathbf{h}_{i})-f(\mathbf{h}_{j})|| =||\mathbf{w} \odot \mathbf{h}_i - \mathbf{w} \odot \mathbf{h}_j||^2
\end{equation}

Before transformation, the learned distance between object $i$ and $j$ is $d_{ij}$ (Eq.~\ref{eq:d_ij}), which can be skewed by noisy and irrelevant features. After transformation, the distance $d_{ij}^{'}$ (Eq.~\ref{eq:d_ij_weighted}) can be more informative for classification tasks. Note the kNN attention pooling (Eq.~\ref{eq:knn_pooling}) is still used after the feature transformation (Eq.~\ref{eq:feature_attention}). The main difference between the feature attention layer and the kNN pooling layer is that the feature attention layer uses element-wise multiplication  (Eq.~\ref{eq:feature_attention}) instead of \textit{affine} layer followed by \textit{ReLU()} (Eq.~\ref{eq:f}) as nonlinear transformation.
Just like skip connections in ResNet \citep{He2016} that can help gradient flow, the feature attention layer can help select important \textit{individual} features much easier than the fully connected layer, and can increase the generalization power of a neural network model in certain cases with very few training examples. 

In addition, for fully connected \textit{affine} layer without weight constraints, the weight can be negative and unbounded. Even if we set non-negativity constraints to the weight, the transformed features are linear combinations of the input features. We cannot directly determine the importance of individual features.
By contrast, the feature attention layer only has parameter $w$ (Eq.~\ref{eq:w}), which directly corresponds to the learned feature weight. Because of the constraint on $w$ (Eq.~\ref{eq:w}), the feature attention layer also learns a weighted Euclidean metric during training.

\section{Experiments}
\subsection{Simulations} \label{sec:simulation}
\begin{figure}[t!]
		\centering
		\begin{subfigure}[b]{0.5\textwidth}
			\centering
			\includegraphics[height=1.65in]{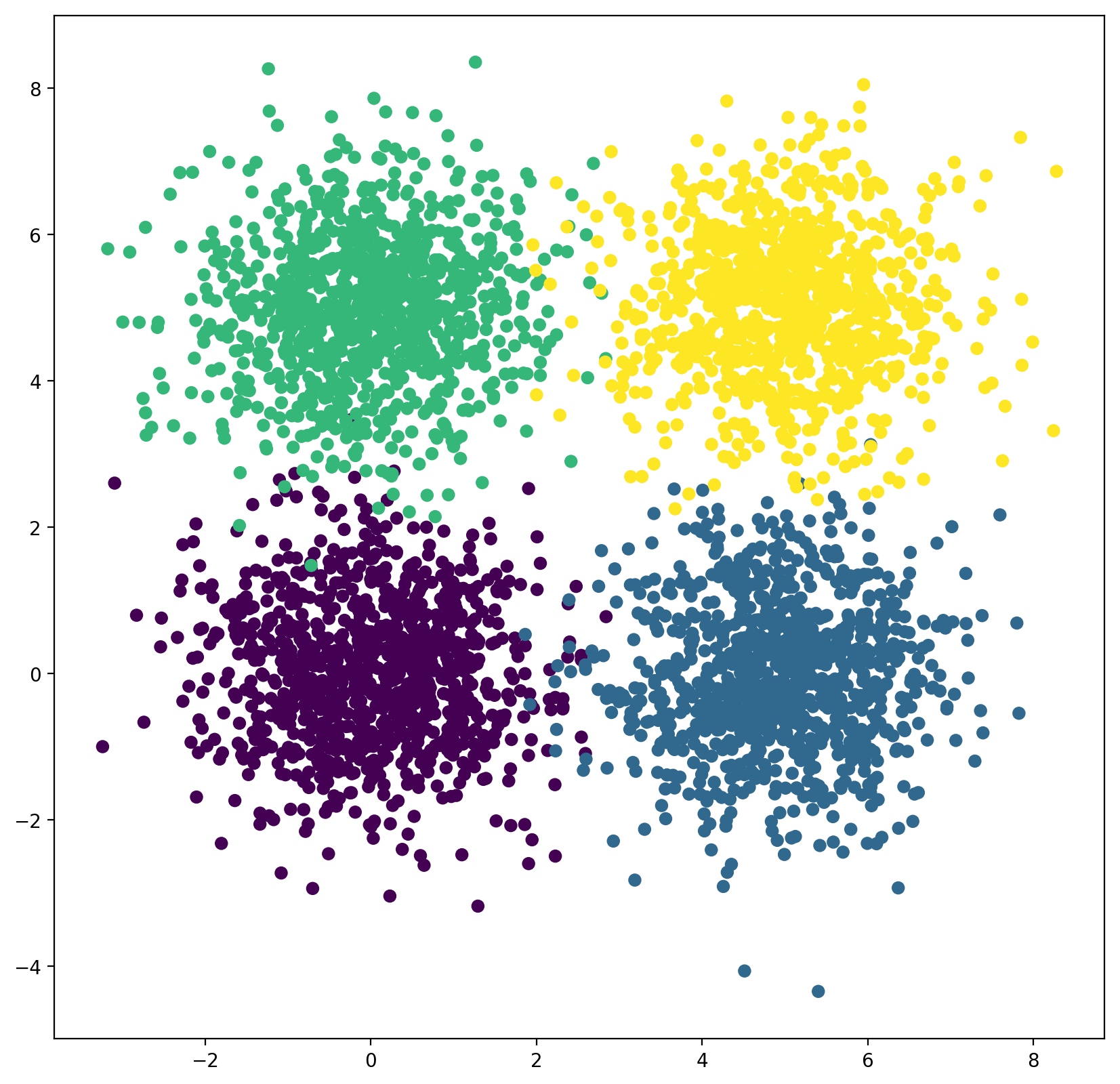}
			\caption{4000 points belonging to four ``natural'' clusters}
			\label{fig:pca_signal}
		\end{subfigure}%
		~ 
		\begin{subfigure}[b]{0.5\textwidth}
			\centering
			\includegraphics[height=1.65in]{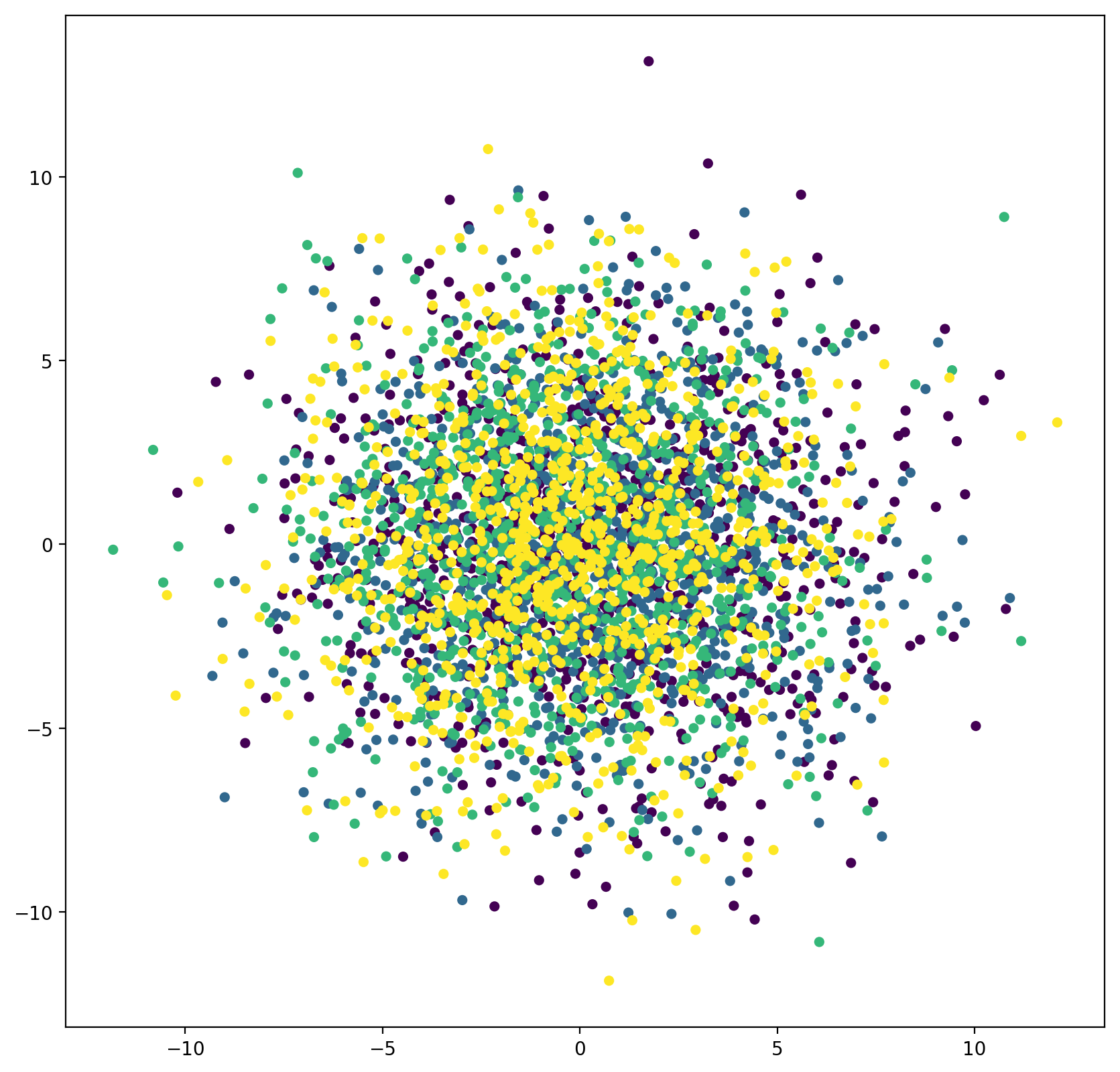}
			\caption{After adding 40-dimensional Gaussian noise}
			\label{fig:pca_noise}
		\end{subfigure}
		\caption{Plots of the true signal and the ``corrupted'' signal}
		\label{fig:pca_raw_data}
	\end{figure}

We sampled 1000 points from each of the four 2-dimensional Gaussian distributions with the same covariance matrix $\Sigma = diag(1,1)$ and four different mean ($\mu = (0,0), (0,5), (5,0), (5,5)$, respectively) as the true signal.
We then appended the true signal with 40-dimensional Gaussian noise with mean $\mu = (2.5, 2.5, \cdots, 2.5)$ and covariance $\Sigma = diag(10, 10, \cdots, 10)$. Thus each point has 42 dimensions, with the first two containing the true signal, and the rest being random noise. With four different colors corresponding to the true cluster assignments (generated from four distributions), we plotted the true signal (i.e., the first two dimensions) in Fig.~\ref{fig:pca_signal} and the ``corrupted'' signal (i.e., 42-dimensional vector) using PCA in Fig.~\ref{fig:pca_noise}. While the true signal forms four ``natural'' clusters (Fig.~\ref{fig:pca_signal}), the corrupted signal is dominated by the added irrelevant features and the clusters are no longer obvious. 
	
%	\begin{figure}[t!]
%		\centering
%		\includegraphics[height=1.63in]{input-sim-data}
%		\caption{Synthetic data: four ``natural'' clusters of points}
%		\label{fig:input-sim-data}
%	\end{figure}

We constructed two models to predict class labels: 

``NeuralNet'': a neural network model with an input layer (42-dimensional), a hidden layer (100 hidden units) and an output layer (4 units corresponding to four classes); 

``AffinityNet'': same as ``NeuralNet'' model except adding one feature attention layer followed by kNN attention pooling after the input layer.
	
\begin{figure}
		\centering
		\begin{subfigure}[t]{0.5\textwidth}
			\centering
			\includegraphics[height=1.65in]{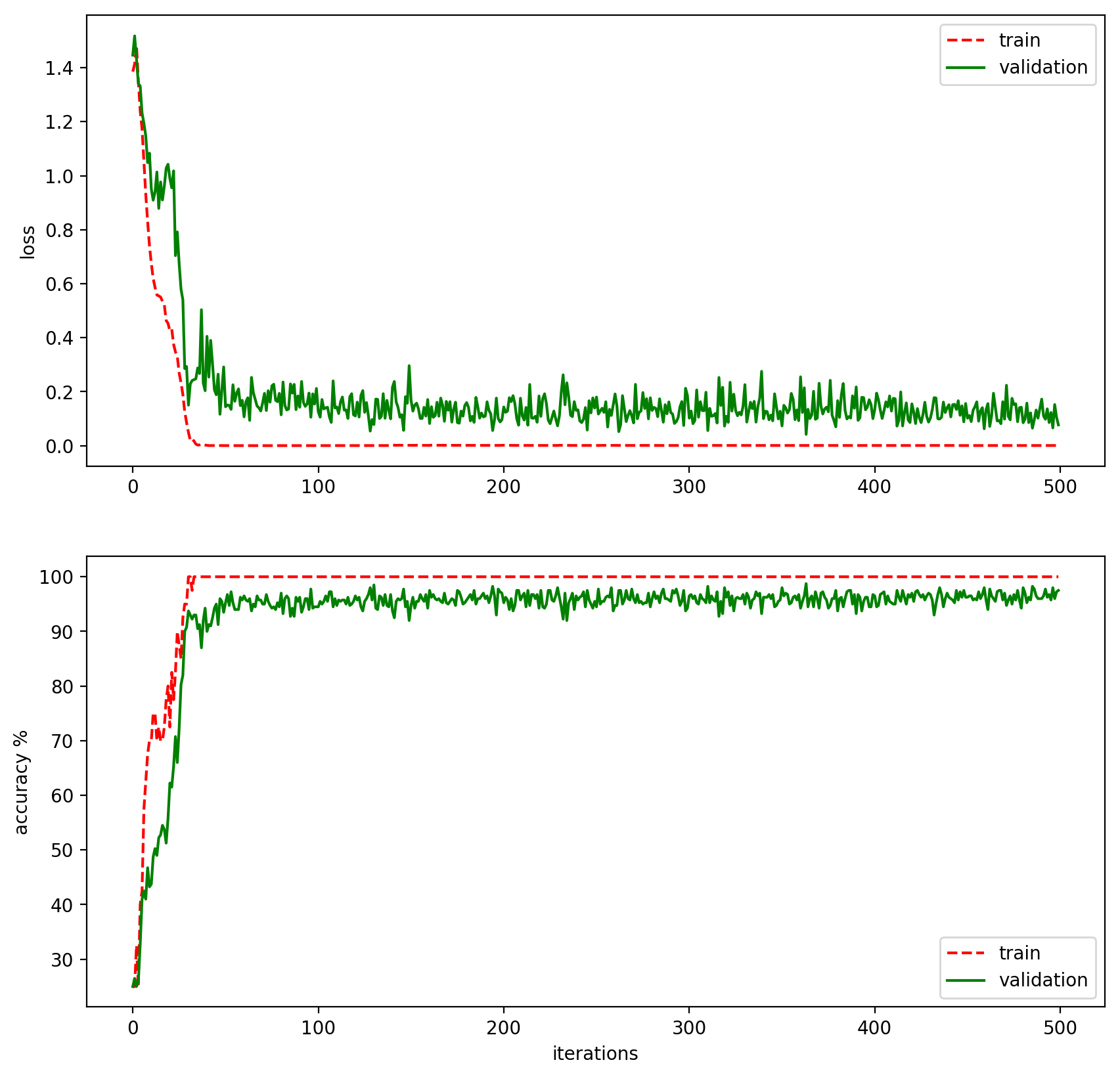}
			\caption{AffinityNet training loss and accuracy}
			\label{fig:sim_loss_acc_affinitynet}
		\end{subfigure}%
		~ 
		\begin{subfigure}[t]{0.5\textwidth}
			\centering
			\includegraphics[height=1.65in]{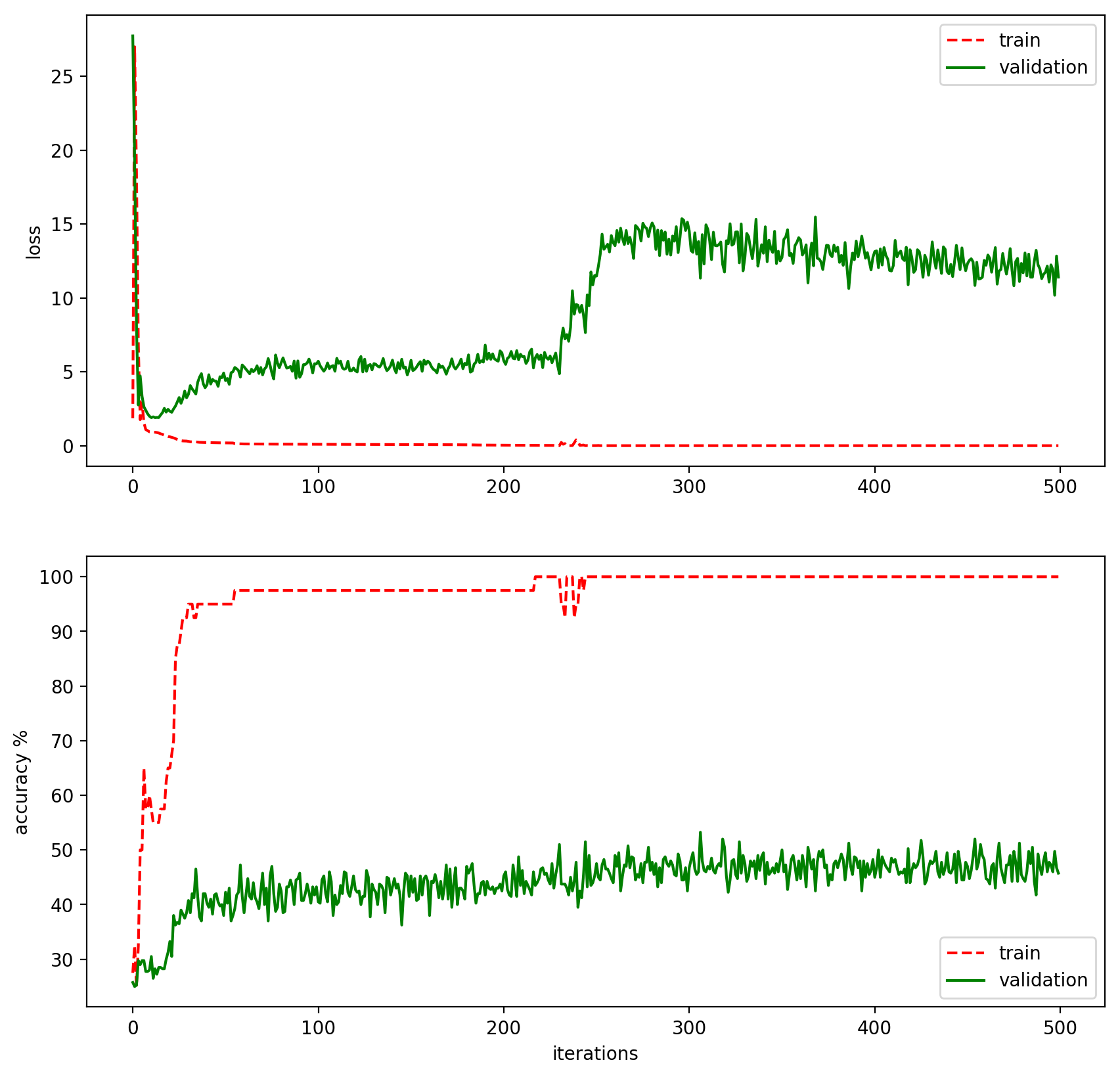}
			\caption{NeuralNet training loss and accuracy}
			\label{fig:sim_loss_acc_nn}
		\end{subfigure}
		~
		\begin{subfigure}[t]{0.5\textwidth}
			\centering
			\includegraphics[height=1.65in]{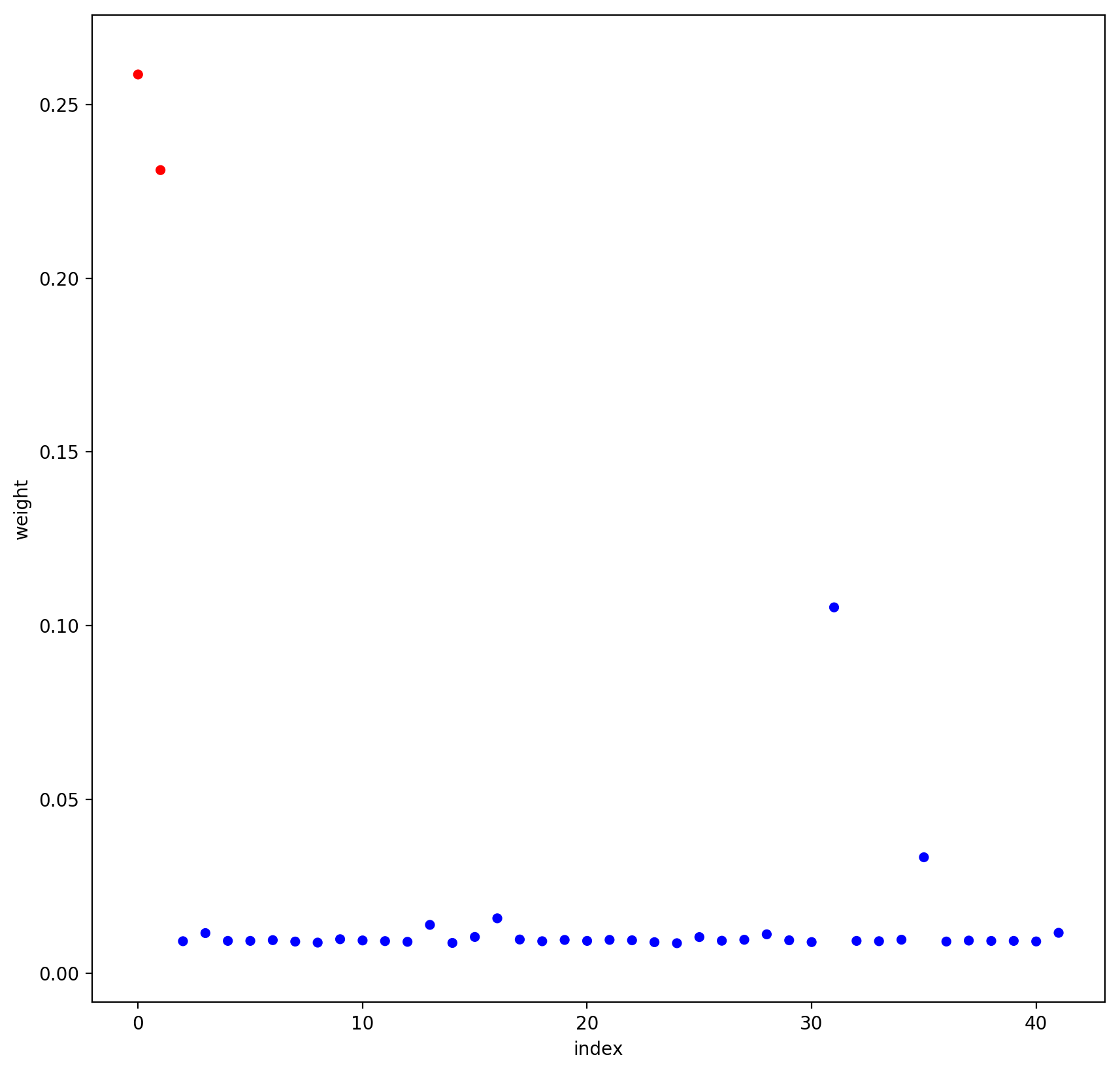}
			\caption{feature weights learned by AffinityNet}
			\label{fig:sim_weight_affinitynet}
		\end{subfigure}%
		~ 
		\begin{subfigure}[t]{0.5\textwidth}
			\centering
			\includegraphics[height=1.65in]{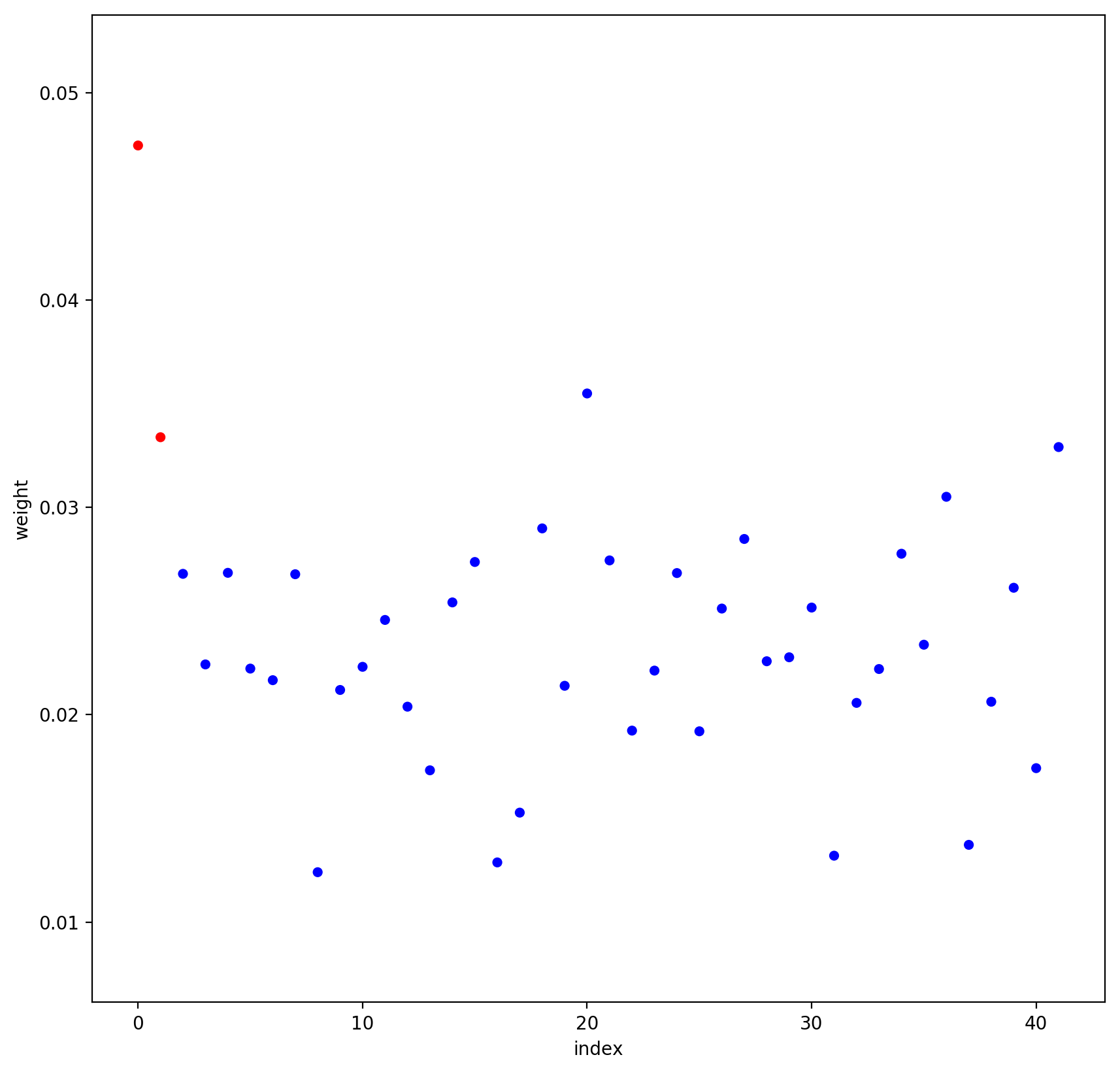}
			\caption{feature weights learned by NeuralNet}
			\label{fig:sim_weight_nn}
		\end{subfigure}
		\caption{Training loss and accuracy and learned feature weights}
		\label{fig:loss_acc_weight}
	\end{figure}

We randomly selected 1\% of data (40 out of 4000 points) for training two models and compared accuracies on the test set. Surprisingly, by only training 1\% of the data, our model with feature attention layer can successfully select the true signal features and achieve 98.2\% accuracy on the test set. By contrast, a plain neural network model only achieved 46.9\% accuracy on test set.
Fig.~\ref{fig:sim_loss_acc_affinitynet} and Fig.~\ref{fig:sim_loss_acc_nn} show the training loss and accuracy curves (the red curves are for training set and the green ones for test set) for ``AffinityNet'' and ``NeuralNet'', respectively. Even though both models achieve 100\% training accuracy within a few iterations, the ``AffinityNet'' model generalizes better than the plain neural network model (there is a big gap between training and test accuracy curves for ``NeuralNet'' model when training data is small).	

Strikingly, the good generalization of our model partly relies on the success of the feature attention layer picking up the true signals from the noise. Fig.~\ref{fig:sim_weight_affinitynet} and Fig.~\ref{fig:sim_weight_nn} shows the learned weights by AffinityNet and NeuralNet for the 42-dimensional input features, with the red dots corresponding to the true signals and blue dots noise. The weights of the true signal are much higher than those noise in AffinityNet, while ``NeuralNet'' did not select the true signal very well.

\subsection{Tumor disease type classification}
Harmonized kidney and uterus cancer gene expression datasets were downloaded from Genomic Data Commons Data Portal (\href{https://portal.gdc.cancer.gov}{https://portal.gdc.cancer.gov}) \citep{Grossman2016}.
Kidney cancer has three disease types, and uterus cancer has two. 
The number of samples from each disease type is summarized in Table~\ref{tbl:sample_info_4types}. Both kidney cancer and uterus cancer have unbalanced classes (i.e., one class has much less samples than the other). We calculated the standard deviation of gene expression values for each gene across samples within a cancer type (i.e., either kidney or uterus) and selected top 1000 most variant gene expression features as the input to our model. We are trying to classify each tumor sample into its disease type for uterus and kidney cancer separately using the gene expression profiles. 
	
	\begin{table}[!t]
		\begin{center}
			\caption{Sample information of four cancer types}
			\label{tbl:sample_info_4types}
			\begin{tabular}{cccc}
				\hline
				Cancer type &  \multicolumn{2}{c}{Disease type} & Total \\
				\hline
				\multirow{3}{*}{kidney} & Chromophobe
				& 65 & \multirow{3}{*}{654}\\
				& Renal Clear Cell Carcinoma & 316 & \\
				& Renal Papillary Cell Carcinoma & 273 & \\
				\hline
				\multirow{2}{*}{Uterus} & Uterine Corpus Endometrial Carcinoma
				& 421 & \multirow{2}{*}{475}\\
				\cline{2-3}
				& Uterine Carcinosarcoma
				& 54 & \\
				\hline
			\end{tabular}
		\end{center}
	\end{table}
	 	
We compared our model (``AffinityNet'') with five other methods: ``NeuralNet'' (conventional deep learning model), ``SVM'', ``Naive Bayes'', ``Random Forest'', and ``Nearest Neighbors'' (kNN).
Our model (``AffinityNet'') consists of 
a feature attention layer, a kNN attention pooling layer (100 hidden units), and a fully connected layer.
For kNN attention pooling layer, we use ``cosine similarity'' kernel and set the number of nearest neighbors $k=2$ (kidney cancer) and $k=3$ (uterus cancer). We have tried other choices of $k$ and the results are similar.
``NeuralNet'' is a two-layer fully connected neural network with the hidden layer having 100 hidden units. For both ``AffinityNet'' and ``NeuralNet'', we use \textit{ReLU()} nonlinear activation in the hidden layer.
Since the input dimension is 1000 (i.e., top 1000 most variant gene expressions), the total parameters of ``NeuralNet'' is 100,403 parameters for kidney cancer with three classes (i.e., disease types), and 100,202 parameters for uterus cancer with two classes.
Our model ``AffinityNet'' has 101,403 parameters and 101,202 parameters for kidney and uterus cancer, respectively. Note our model only has 1000 more parameters than ``NeuralNet'' to facilitate fair comparisons.
We do not use more layers in the neural network models because there are only several hundred samples to train, and larger models are more likely to overfit.
We used the implementation from scikit-learn (\href{http://scikit-learn.org}{http://scikit-learn.org}) for ``Naive Bayes'', ``SVM'',  ``Nearest Neighbors'', and ``Random Forest'' with default settings.
	
We progressively increased the training portion from 1\% to 70\% (i.e., 1\%, 10\%, 20\%, 30\%, 40\%, 50\%, 60\% and 70\%), and reported the adjusted mutual information (AMI) on the test set (Table~\ref{tbl:kidney_nmi_mean} and Table~\ref{tbl:uterus_nmi_mean}). AMI is an adjustment of the Mutual Information (MI) score to account for chance, which is suitable to measure the performance of clustering and classification with multiple unbalanced classes (AUC is a similar metric but is mainly suitable for binary classification).
	
We ran experiments 20 times with different random seeds to generate different training and test sets. For each run, the training and test set for all six methods are identical. We reported the mean AMI scores for the top 10 runs (results depending on the few selected training examples and other randomness) for all methods in Table~\ref{tbl:kidney_nmi_mean} and  Table.~\ref{tbl:uterus_nmi_mean}.
	
For both cancer types, our model clearly outperformed all other models, especially when training portion is small. For example, when trained on only 1\% of the data, our model achieved AMI=0.84 for kidney cancer and AMI=0.62 for uterus cancer  (Table~\ref{tbl:kidney_nmi_mean} and Table~\ref{tbl:uterus_nmi_mean}), while other methods performed badly with few training examples. This suggests our model is highly data efficient.
One reason for this is that kNN attention pooling layer is in a sense performing ``clustering'' during training, and it is less likely to overfit a small number of training examples. The input of kNN attention pooling layer can contain not only labeled training examples but also unlabeled examples. It performs semi-supervised learning with a few labeled examples as a guidance for finding ``clusters'' among all the data points. ``NeuralNet'' and other methods do not perform well with few labeled training examples because they tend to overfit the training set. As more training data is available, other methods including ``NeuralNet'' are improving rapidly. In this case, ``NeuralNet'' model does not outperform traditional machine learning techniques such as ``SVM'' because the dataset is quite small and the power of deep learning is manifested only when large amounts of data is available.
	
	%\begin{figure*}[t!]
	%	\centering
	%	\includegraphics[width=\textwidth]{kidney_test_nmi}
	%	\caption{Adjusted Mutual Information on test set for kidney cancer}
	%	\label{fig:kidney_test_nmi}
	%\end{figure*}

	%In Figure~\ref{fig:kidney_test_nmi}, 
In Table~\ref{tbl:kidney_nmi_mean},
note that for kidney cancer, unlike other methods, our model does not improve with more training data, partly because there are a few very hard cases in kidney cancer dataset, while all other cases are almost linearly separable. Our model can easily pick up the linearly separable clusters with only a few training examples, but it is hard to separate very hard cases even when more training data is available.
Uterus cancer dataset is highly unbalanced with one class being much smaller than the other, and thus it is much harder to achieve high adjusted mutual information (AMI). As shown in Table~\ref{tbl:uterus_nmi_mean}, ``AffinityNet'' achieved AMI$=0.62$ when trained on approximately 1\% of the data (i.e., randomly chosen 1 samples from disease Uterine Carcinosarcoma and 4 samples from the other disease type), and performed significantly better than other models even as the training portion increased to 70\%. This suggests ``AffinityNet'' works well on highly unbalanced data, while other methods seems to be inadequate.	

	\begin{table}[t!]
		\begin{center}
			\caption{Adjusted Mutual Information on test set for kidney cancer}
			\label{tbl:kidney_nmi_mean}
			\begin{tabular}{ccccccccc}
				\hline
				\toprule
				& \multicolumn{8}{c}{Train portion} \\
				\cmidrule(lr){2-9}
				Method &0.01 & 0.1 & 0.2 & 0.3 & 0.4 & 0.5 & 0.6 & 0.7\\\hline
				AffinityNet & \textbf{0.84} & \textbf{0.87} & \textbf{0.86} & \textbf{0.85} & \textbf{0.85} & \textbf{0.85} & \textbf{0.86} & \textbf{0.85} \\\hline
				NeuralNet & 0.70 & 0.76 & 0.77 & 0.78 & 0.78 & 0.80 & 0.80 & 0.81 \\\hline
				SVM & 0.70 & 0.77 & 0.78 & 0.79 & 0.80 & 0.81 & 0.82 & 0.83 \\\hline
				Naive Bayes & 0.25 & 0.59 & 0.71 & 0.76 & 0.78 & 0.80 & 0.81 & 0.83 \\\hline
				Random Forest & 0.36 & 0.61 & 0.67 & 0.68 & 0.71 & 0.72 & 0.73 & 0.75
				\\\hline
			\end{tabular}
		\end{center}
	\end{table}
	
	%\begin{table}[t!]
	%	\begin{center}
	%		\caption{Adjusted Mutual Information on test set for uterus cancer}
	%		\label{tbl:uterus_nmi_max}
	%		\begin{tabular}{|c|cccccccc|}
	%			\hline
	%			\backslashbox{method}{train\_portion} &0.01 & 0.1 & 0.2 & 0.3 & 0.4 & 0.5 & 0.6 & 0.7\\\hline
	%			Affinity Net & 0.82 & 0.89 & 0.87 & 0.90 & 0.94 & 0.92 & 1.00 & 1.00 \\\hline
	%			Neural Net & 0.50 & 0.59 & 0.66 & 0.69 & 0.71 & 0.78 & 0.80 & 0.81 \\\hline
	%			Linear SVM & 0.54 & 0.62 & 0.66 & 0.69 & 0.75 & 0.86 & 0.91 & 0.88 \\\hline
	%			Naive Bayes & 0.00 & 0.47 & 0.65 & 0.65 & 0.64 & 0.69 & 0.69 & 0.64 \\\hline
	%			Random Forest & 0.16 & 0.14 & 0.17 & 0.17 & 0.33 & 0.22 & 0.31 & 0.33
	%			\\\hline
	%		\end{tabular}
	%	\end{center}
	%\end{table}

	\begin{table}[t!]
		\begin{center}
			\caption{Adjusted Mutual Information on test set for uterus cancer}
			\label{tbl:uterus_nmi_mean}
			%		\hspace*{-1cm}
			\begin{tabular}{ccccccccc}
				\hline
				\toprule
				& \multicolumn{8}{c}{Train portion} \\
				\cmidrule(lr){2-9}
				Method &0.01 & 0.1 & 0.2 & 0.3 & 0.4 & 0.5 & 0.6 & 0.7\\\hline
				AffinityNet & \textbf{0.62} & \textbf{0.66} & \textbf{0.76} & \textbf{0.84} & \textbf{0.85} & \textbf{0.84} & \textbf{0.85} & \textbf{0.86} \\\hline
				NeuralNet & 0.41 & 0.52 & 0.59 & 0.61 & 0.63 & 0.68 & 0.69 & 0.73 \\\hline
				SVM & 0.43 & 0.54 & 0.60 & 0.62 & 0.65 & 0.70 & 0.71 & 0.70 \\\hline
				Naive Bayes & 0.00 & 0.28 & 0.62 & 0.58 & 0.55 & 0.58 & 0.57 & 0.55 \\\hline
				Random Forest & 0.03 & 0.06 & 0.10 & 0.12 & 0.18 & 0.14 & 0.19 & 0.20
				\\\hline
			\end{tabular}
			%	\hspace*{-1cm}
		\end{center}
	\end{table}

\subsection{Semi-supervised clustering}
Cancer patient clustering and disease subtype discovery is very challenging because of the small sample size and lack of enough training examples with groundtruth labels. 
If we can obtain label information for a few samples, we can use ``AffinityNet'' for semi-supervised clustering \citep{Weston2012}.
While other methods such as SVM do not produce explicit feature representations, both ``AffinityNet'' and ``NeuralNet'' can learn a new feature representation through multiple nonlinear transformations. For a classification model, the new feature representation is usually fed into a linear classifier. We can train our model with a few labeled examples, use the learned model to generate the transformed feature representations for all data points, and then perform clustering using the transformed features.
	
For ``AffinityNet'', we can use all the data points during training with kNN attention pooling, but only backpropagate on labeled training examples. We get the learned new representations for all the data points once the training process is finished. For conventional neural network models, since each data point is independently trained, we only use labeled examples during training. After training, we have to use the learned model to generate new feature representations for all the data points.
In order to evaluate the quality of the learned feature representations with a few training examples, we performed clustering using these transformed features and using the original features, and compared them with groundtruth class labels.
	
	\begin{figure}[t!]
		\centering
		\includegraphics[width=.8\textwidth]{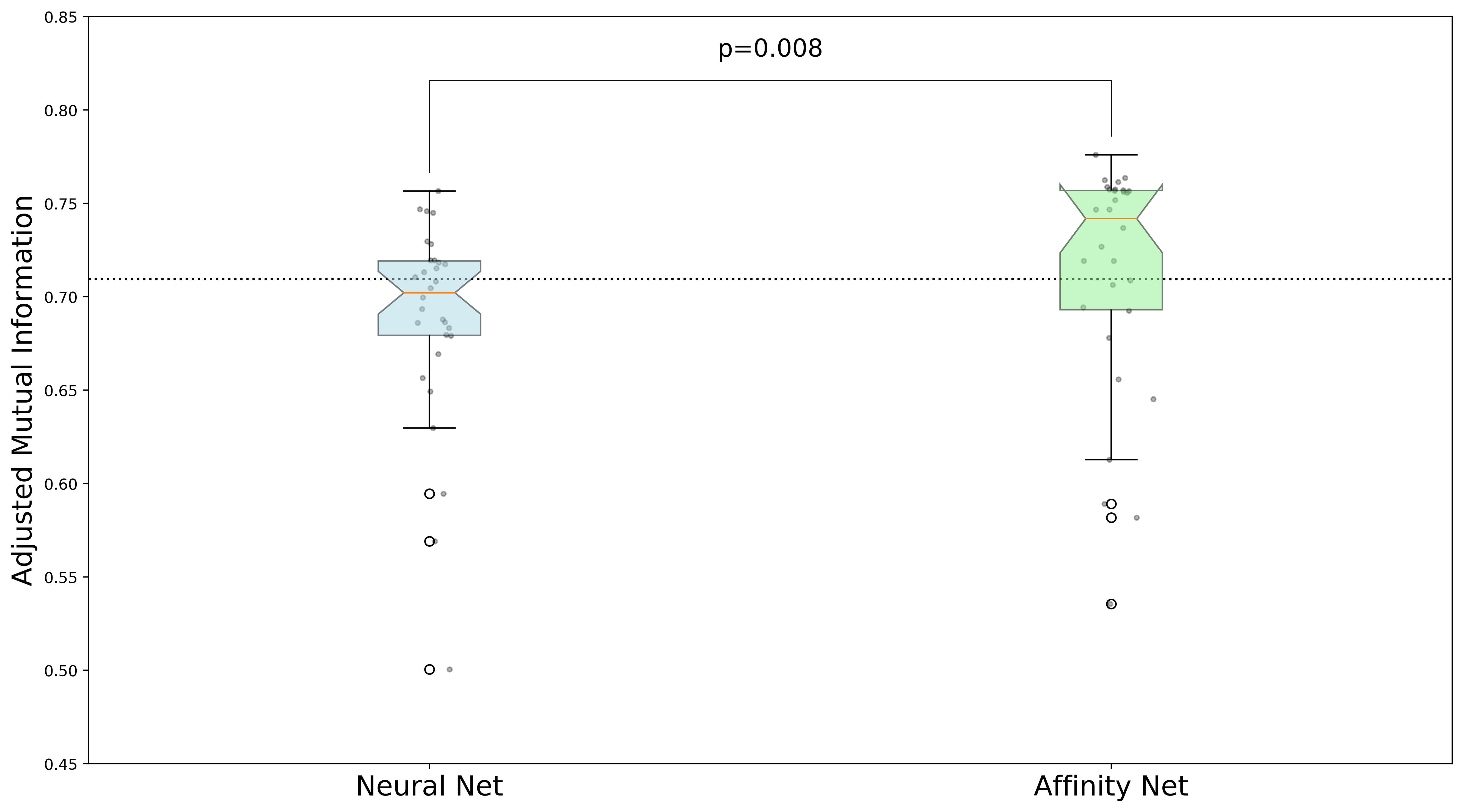}
		\caption{Adjusted Mutual Informtion achieved by semi-supervised clustering using ``Neural Net'' and  ``Affinity Net'' for kidney cancer}
		\label{fig:semi_clu_boxplot}
	\end{figure}
	
We compared the performance using  ``AffinityNet'' and ``NeuralNet'' on kidney data set as it has more samples.
We randomly selected 1\% of data for training, and ran experiments 30 times. After training, we performed spectral clustering on transformed patient-feature matrix.
Fig.~\ref{fig:semi_clu_boxplot} shows the adjusted mutual information scores for all  the 30 runs using ``AffinityNet'' and ``NeuralNet''. We also performed spectral clustering on the original patient-feature matrix as a baseline method (AMI = 0.71, blue dotted line in the figure).
Our model outperformed the ``NeuralNetwork'' model ($p=0.008$, Wilcoxon signed rank test) and the baseline (the ``Neural Network'' model is slightly below the baseline because it probably had overfitted the training examples). While both ``NeuralNet'' and ``AffinityNet'' have approximately the same number of model parameters, only ``AffinityNet'' can learn a good feature transformation by facilitating semi-supervised few-shot learning with feature attention and kNN attention pooling layers.
	
\subsection{Combine with Cox model for survival analysis}
For many cancer genomics studies, cancer subtype information is not known, but patient survival information is available. We replaced the last layer (i.e., linear classifier) in the model (as shown in Fig.~\ref{fig:model-overview}) with a regression layer following the Cox proportional hazards model \citep{Mobadersany2018,Fox2002}. We used backpropagation to learn model parameters that maximize partial likelihood in the Cox model.
	
\begin{figure}
		\centering
		\includegraphics[width=0.8\textwidth]{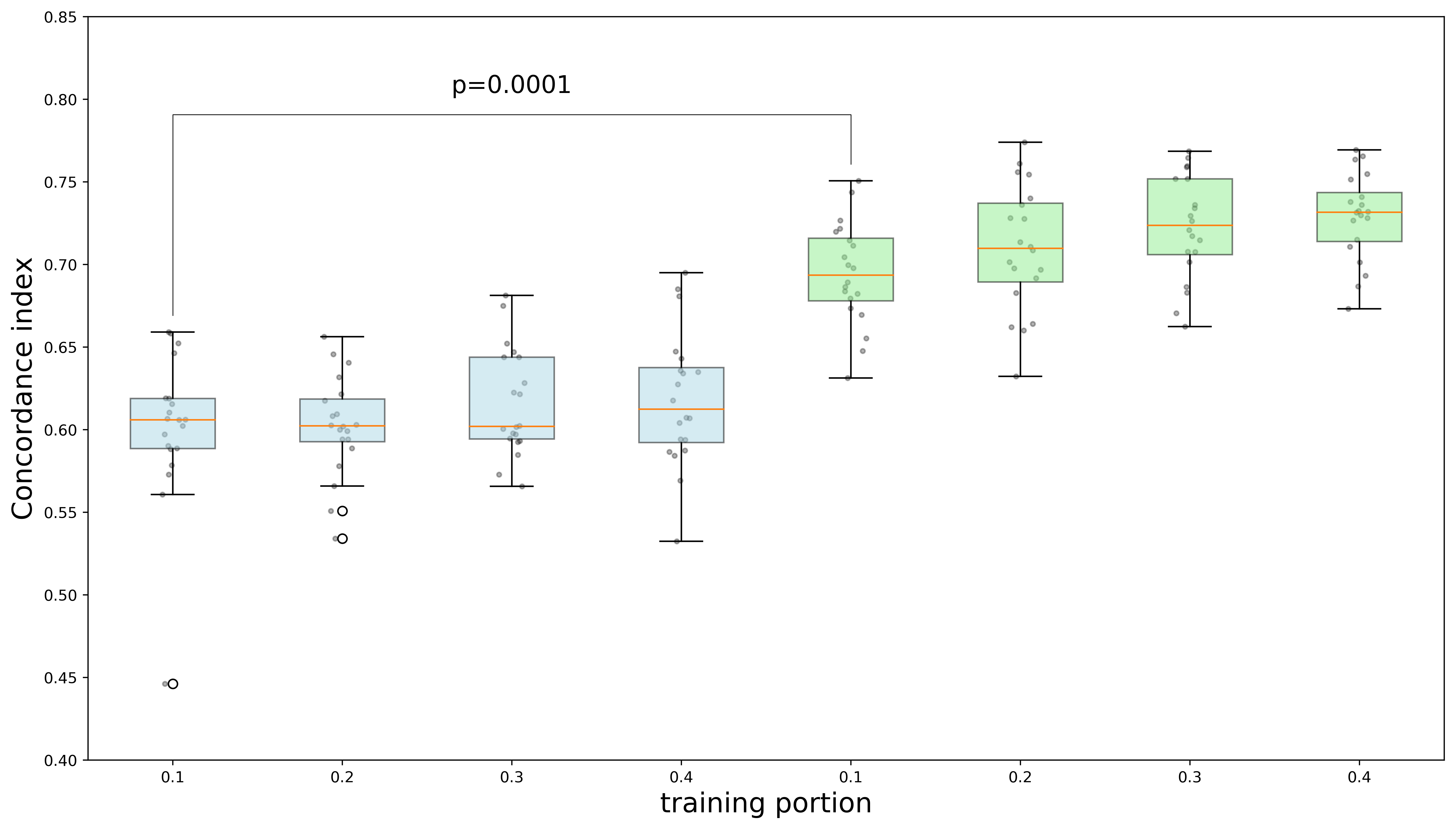}
		\caption{Concordance index achieved by ``Affinity Net'' and baseline Cox model for kidney cancer}\label{fig:c_index_boxplot}
	\end{figure}

We performed experiments on kidney cancer dataset that has more than 600 samples. We progressively increased the training portion from 10\% to 40\%. We used 30\% of data as validation and the remaining as test set.
As a baseline method, we used age, gender and known disease types as covariates to fit a Cox model.
We ran experiments 20 times with random seeds, and reported the concordance index on the test set for both our model and the baseline Cox model (Fig.~\ref{fig:c_index_boxplot}).
	
In Fig.~\ref{fig:c_index_boxplot}, the light blue boxplots on the left side correspond to the results from the baseline method (i.e., the Cox model on age, gender and disease types), while the light green ones correspond to that from our model. The reported p-value between our model and the baseline method for training 10\% data was calculated using Wilcoxon signed rank test. Our model outperformed the baseline model by a significant margin (Table.~\ref{tbl:concordance_score} shows the mean concordance index in different settings).
	
	\begin{table}[t!]
		\begin{center}
			\caption{Mean concordance scores for kidney cancer}
			\label{tbl:concordance_score}
			\begin{tabular}{ccccc}
				\hline
				\toprule
				& \multicolumn{4}{c}{Train portion} \\
				\cmidrule(lr){2-5}
				Method &0.1 & 0.2 & 0.3 & 0.4\\\hline
				Baseline Cox Model & 0.601 & 0.602 & 0.616 & 0.618 \\\hline
				Affinity Net & 0.694 & 0.710 & 0.723 & 0.729
				\\\hline
			\end{tabular}
		\end{center}
	\end{table}

\begin{figure}[!t]
		\centering
		\includegraphics[width=.8\textwidth]{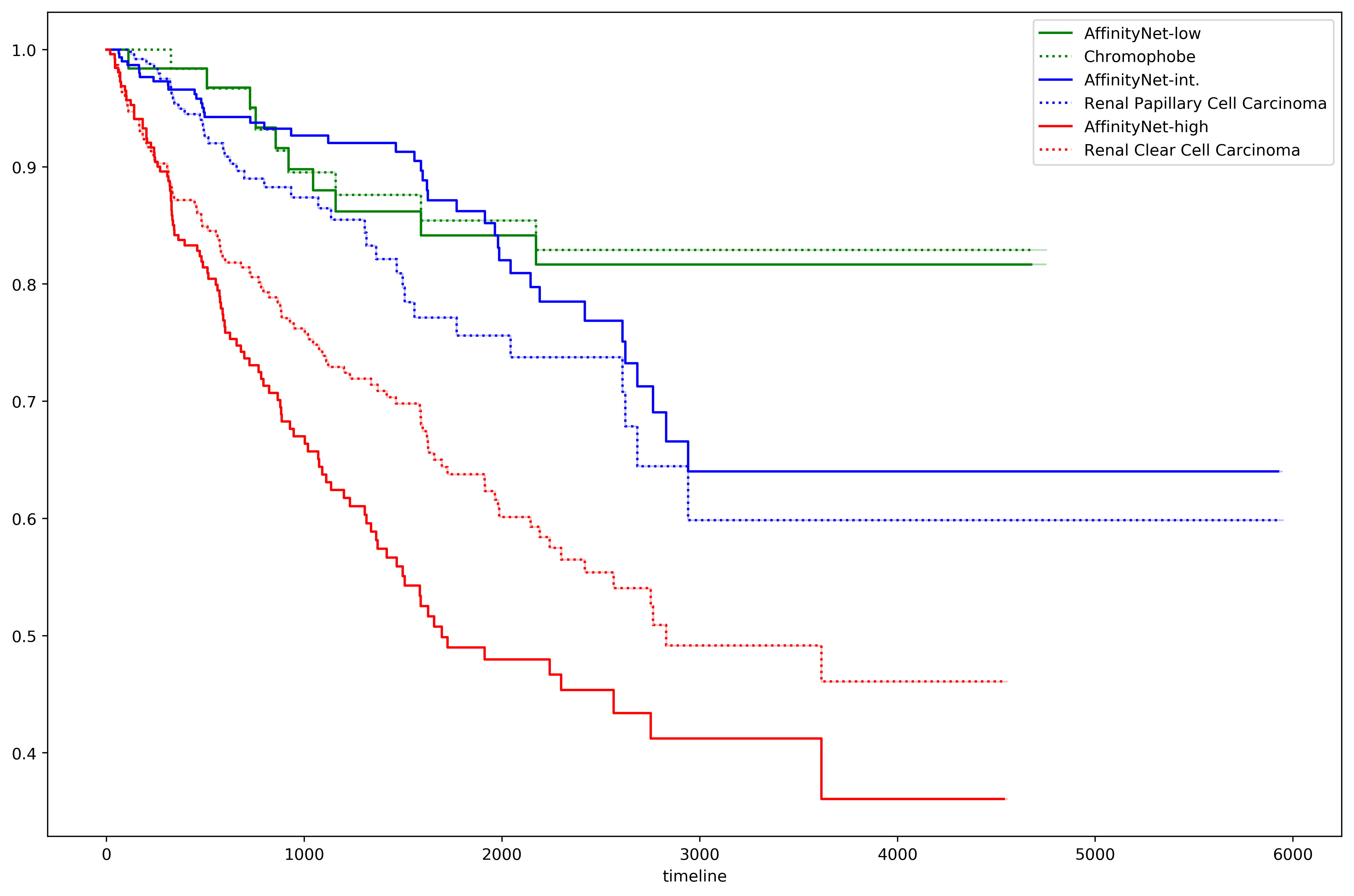}
		\caption{Kaplan Meier plot comparing kidney cancer disease types and AffinityNet predictions}\label{fig:km-plot}
	\end{figure}

There are three disease types of kidney cancer. We used our best model trained on 10\% of the data to calculate the hazard rates for all kidney cancer patients in the dataset, and split them into three groups with low, intermediate, and high hazard rates. The proportions of the three groups are the same as the three disease types. Fig.~\ref{fig:km-plot} shows the Kaplan Meier plot for both three known disease types (dotted line) and three groups based on the predicted hazard rates (AffinityNet-low, AffinityNet-int., and AffinityNet-high in the figure).
The p-value of log rank test of our predicted groups is $p=6.7\times 10^{-16}$, while the p-value of log rank test for three known disease types is $p=1.5\times 10^{-5}$, indicating our model can better separate patients with a different survival time.
	
\section{Discussion and Conclusion}
	
Deep learning has achieved great success in computer vision, natural language processing, and speech recognition, where features (e.g., pixels, words, and audio signals) are well structured and a large amount of training data is available. However, in biomedical research, the training sample size is usually small while the feature dimension is very high, where deep learning models tend to overfit the training data but fail to generalize.
To alleviate this problem in the patient clustering/classification related tasks, we propose the AffinityNet model that contains stacked feature attention and kNN attention pooling layers to facilitate semi-supervised few-shot learning. 
	
Regardless of whether a graph is given or not, kNN attention pooling layer can use attention kernels to calculate dynamic affinity graphs during training. The affinity graphs are used for selecting k-nearest neighbors for attention-based pooling. kNN attention pooling layers essentially add a ``clustering'' operation (``forcing'' similar objects to have similar representations through attention-based pooling) after the nonlinear feature transformations, which can serve as an implicit regularizer for classification-related tasks.
kNN attention pooling layers can be plugged into a deep learning model as a basic building block just like convolutional layers. With multi-view data, we can first use a few kNN attention pooling layers to process each view separately to learn a high-level representation for each view, and then combine all the views with their high-level feature representations (by concatenating them together or adding them up) and apply kNN attention pooling again to the combined view.
Feature attention layer is a simple special case of kNN attention pooling layer. It is useful for selecting important individual input features automatically with a normalized non-negative weight learned for each feature. 

Building upon stacked feature attention and kNN pooling layers, our AffinityNet model is more effective for semi-supervised few-shot learning than conventional deep learning models. 
We have conducted extensive experiments using AffinityNet on two cancer genomics datasets and achieved satisfactory results.

AffinityNet alleviates the problem of lack of a sufficient amount of labeled training data by utilizing unlabeled data with kNN attention pooling, and can be used to analyze a large bulk of cancer genomics data for patient clustering and disease subtype discovery. Future work may focus on designing deep learning modules that can incorporate biological knowledge for various tasks.
	
%	\section*{Funding}
%	
%	This work was supported in part by the US National Science Foundation under grants NSF IIS-1218393 and IIS-1514204. Any opinions, findings, and conclusions or recommendations expressed in this material are those of the author(s) and do not necessarily reflect the views of the National Science Foundation. \vspace*{-12pt}

%\bibliography{references}
%\bibliographystyle{icml2018}

\end{document}